\documentclass[11pt,letterpaper]{article}
\usepackage{naaclhlt2016}
\usepackage{times}
\usepackage{latexsym}

\usepackage{graphicx}
\usepackage{CJK}
\usepackage{enumitem}
\usepackage{multirow}
\usepackage{array}
\usepackage{color,soul} 
\usepackage{url}
\usepackage{amsmath}
\usepackage{amsfonts}

\usepackage{balance}

\naaclfinalcopy 
\def\naaclpaperid{***} 

\title{A Novel Approach to Dropped Pronoun Translation}

\def\fndaff{$^\dagger$}
\def\sstaff{$^\ddag$}
\author{Longyue Wang\fndaff ~~~~ Zhaopeng Tu\sstaff ~~~~ Xiaojun Zhang\fndaff ~~~~ Hang Li\sstaff ~~~~ Andy Way\fndaff ~~~~ Qun Liu\fndaff\\
\fndaff ADAPT Centre, School of Computing, Dublin City University, Ireland \\ 
{\tt \{lwang, xzhang, away, qliu\}@computing.dcu.ie}\\
\sstaff Noah's Ark Lab, Huawei Technologies, China\\
{\tt \{tu.zhaopeng, hangli.hl\}@huawei.com}}

\date{}

\begin{document}

\maketitle

\begin{abstract}
Dropped Pronouns (DP) in which pronouns are frequently dropped in the source language but should be retained in the target language are challenge in machine translation. In response to this problem, we propose a semi-supervised approach to recall possibly missing pronouns in the translation. Firstly, we build training data for DP generation in which the DPs are automatically labelled according to the alignment information from a parallel corpus. Secondly, we build a deep learning-based DP generator for input sentences in decoding when no corresponding references exist. More specifically, the generation is two-phase: (1) DP position detection, which is modeled as a sequential labelling task with recurrent neural networks; and (2) DP prediction, which employs a multilayer perceptron with rich features. Finally, we integrate the above outputs into our translation system to recall missing pronouns by both extracting rules from the DP-labelled training data and translating the DP-generated input sentences. Experimental results show that our approach achieves a significant improvement of 1.58 BLEU points in translation performance with 66\% F-score for DP generation accuracy.
\end{abstract}

\section{Introduction}
\begin{CJK}{UTF8}{gbsn}

\begin{figure}[t]
\graphicspath{ {figures/} }
\centering
\includegraphics[width=0.45\textwidth]{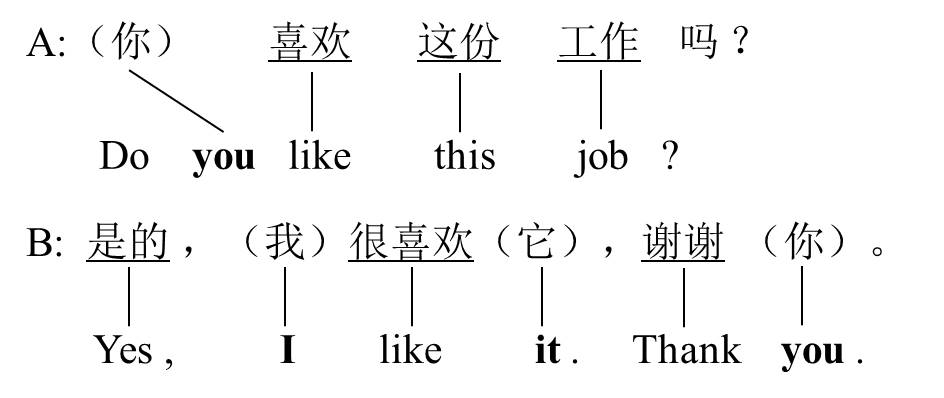}
\caption{Examples of dropped pronouns in a parallel dialogue corpus. The Chinese pronouns in brackets are dropped.}
\label{fig1}
\end{figure}

In pro-drop languages, certain classes of pronouns can be omitted to make the sentence compact yet comprehensible when the identity of the pronouns can be inferred from the context~\cite{yang2015recovering}. 
Figure \ref{fig1} shows an example, in which Chinese is a pro-drop language~\cite{Huang:1984:LI}, while English is not~\cite{haspelmath2001european}.
On the Chinese side, the subject pronouns \{你 (\textit{you}), 我 (\textit{I})\} and the object pronouns \{它 (\textit{it}), 你 (\textit{you})\} are omitted in the dialogue between Speakers $A$ and $B$. These omissions may not be problems for humans since people can easily recall the missing pronouns from the context. However, this poses difficulties for Statistical Machine Translation (SMT) from pro-drop languages (e.g. Chinese) to non-pro-drop languages (e.g. English), since translation of such missing pronouns cannot be normally reproduced.
Generally, this phenomenon is more common in informal genres such as dialogues and conversations than others~\cite{yang2015recovering}. We also validated this finding by analysing a large Chinese--English dialogue corpus which consists of 1M sentence pairs extracted from movie and TV episode subtitles. We found that there are 6.5M Chinese pronouns and 9.4M English pronouns, which shows that more than 2.9 million Chinese pronouns are missing.

In response to this problem, we propose to find a general and replicable way to improve translation quality. 
The main challenge of this research is that training data for DP generation are scarce. Most works either apply manual annotation~\cite{yang2015recovering} or use existing but small-scale resources such as the Penn Treebank~\cite{chung2010effects,xiang2013enlisting}. In contrast, we employ an unsupervised approach to automatically build a large-scale training corpus for DP generation using alignment information from parallel corpora. 
The idea is that parallel corpora available in SMT can be used to project the missing pronouns from the target side (i.e. non-pro-drop language) to the source side (i.e. pro-drop language).
To this end, we propose a simple but effective method: a bi-directional search algorithm with Language Model (LM) scoring.

After building the training data for DP generation, we apply a supervised approach to build our DP generator. We divide the DP generation task into two phases: {\em DP detection} (from which position a pronoun is dropped), and {\em DP prediction} (which pronoun is dropped). Due to the powerful capacity of feature learning and representation learning, we model the DP detection problem as sequential labelling with Recurrent Neural Networks (RNNs) and model the prediction problem as classification with Multi-Layer Perceptron (MLP) using features at various levels: from lexical, through contextual, to syntax.

Finally, we try to improve the translation of missing pronouns by explicitly recalling DPs for both parallel data and monolingual input sentences. More specifically, we extract an additional rule table from the DP-inserted parallel corpus to produce a ``pronoun-complete'' translation model.
In addition, we pre-process the input sentences by inserting possible DPs via the DP generation model. This makes the input sentences more consistent with the additional pronoun-complete rule table.
To alleviate the propagation of DP prediction errors, we feed the translation system $N$-best prediction results via confusion network decoding~\cite{rosti2007combining}.

To validate the effect of the proposed approach, we carried out experiments on a Chinese--English translation task.
Experimental results on a large-scale subtitle corpus show that our approach improves translation performance by 0.61 BLEU points~\cite{Papineni:2002} using the additional translation model trained on the DP-inserted corpus. Working together with DP-generated input sentences achieves a further improvement of nearly 1.0 BLEU point. Furthermore, translation performance with $N$-best integration is much better than its 1-best counterpart (i.e. +0.84 BLEU points).

Generally, the contributions of this paper include the following: 
\begin{itemize}[noitemsep,topsep=2pt]
\setlength\itemsep{0.3em}
\item We propose an automatic method to build a large-scale DP training corpus. Given that the DPs are annotated in the parallel corpus, models trained on this data are more appropriate to the translation task;
\item Benefiting from representation learning, our deep learning-based generation models are able to avoid ignore the complex feature-engineering work while still yielding encouraging results;
\item To decrease the negative effects on translation caused by inserting incorrect DPs, we force the SMT system to arbitrate between multiple ambiguous hypotheses from the DP predictions.
\end{itemize}

The rest of the paper is organized as follows. In Section 2, we describe our approaches to building the DP corpus, DP generator and SMT integration. Related work is described in Section 3. The experimental results for both the DP generator and translation are reported in Section 4. Section 5 analyses some real examples  which is followed by our conclusion in Section 6.
\end{CJK}

\section{Methodology}

\begin{figure*}[t]
\centering
\graphicspath{ {figures/} }
\includegraphics[width=0.8\textwidth]{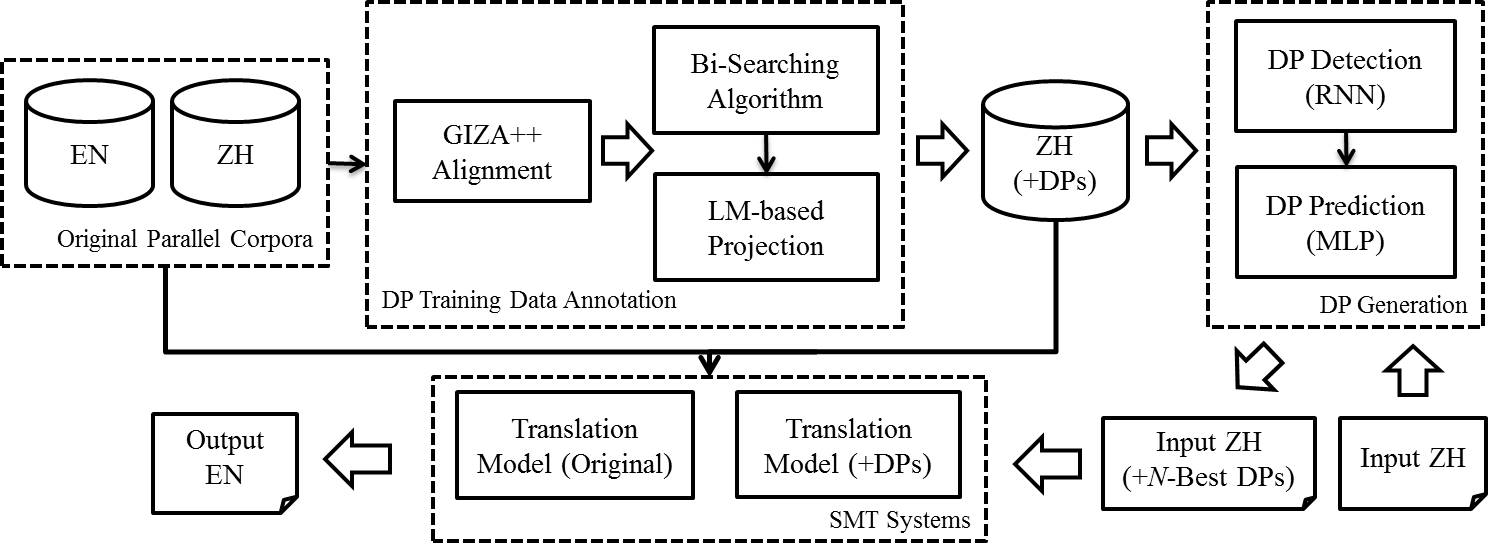}
\caption{Architecture of proposed method.}
\label{fig:2}
\end{figure*}

The architecture of our proposed method is shown in Figure~\ref{fig:2}, which can be divided into three phases: DP corpus annotation, DP generation, and SMT integration.

\subsection{DP Training Corpus Annotation}
\begin{CJK}{UTF8}{gbsn}
We propose an approach to automatically annotate DPs by utilizing alignment information. Given a parallel corpus, we first use an unsupervised word alignment method~\cite{Och:2003,tu2012combining} to produce a word alignment. From observing of the alignment matrix, we found it is possible to detect DPs by projecting misaligned pronouns from the non-pro-drop target side (English) to the pro-drop source side (Chinese).
In this work, we focus on nominative and accusative pronouns including personal, possessive and reflexive instances, as listed in Table~\ref{tab:1}.

\begin{table}[t]
\small
\centering
\begin{tabular}{m{1.5cm}|m{5.2cm}}
\hline
\bf Category & \bf Pronouns  \\\hline

Subjective Personal & 我 (\textit{I}), 我们 (\textit{we}), 你/你们 (\textit{you}), 他 (\textit{he}), 她 (\textit{she}), 它 (\textit{it}), 他们/她们/它们 (\textit{they}).\\\hline

Objective Personal & 我 (\textit{me}), 我们 (\textit{us}), 你/你们 (\textit{you}), 他 (\textit{him}), 她 (\textit{her}), 它 (\textit{it}), 她们/他们/它们 (\textit{them}).\\\hline

Possessive & 我的 (\textit{my}), 我们的 (\textit{our}), 你的/你们的 (\textit{your}), 他的 (\textit{his}), 她的 (\textit{her}), 它的 (\textit{its}), 他们的/她们的/它们的 (\textit{their}).\\\hline

Objective Possessive & 我的 (\textit{mine}), 我们的 (\textit{ours}), 你的/你们的 (\textit{yours}), 他的 (\textit{his}), 她的 (\textit{hers}), 它的 (\textit{its}), 她们的/他们的/它们的 (\textit{theirs}).\\\hline

Reflexive & 我自己 (\textit{myself}), 我们自己 (\textit{ourselves}), 你自己 (\textit{yourself}), 你们自己 (\textit{yourselves}), 他自己 (\textit{himself}), 她自己 (\textit{herself}), 它自己 (\textit{itself}),  他们自己/她们自己/它们自己 (\textit{themselves}). \\\hline

\end{tabular}
\caption{Pronouns and their categories.}\label{tab:1}
\end{table}


\begin{figure}[t]
\centering
\graphicspath{ {figures/} }
\includegraphics[width=0.4\textwidth]{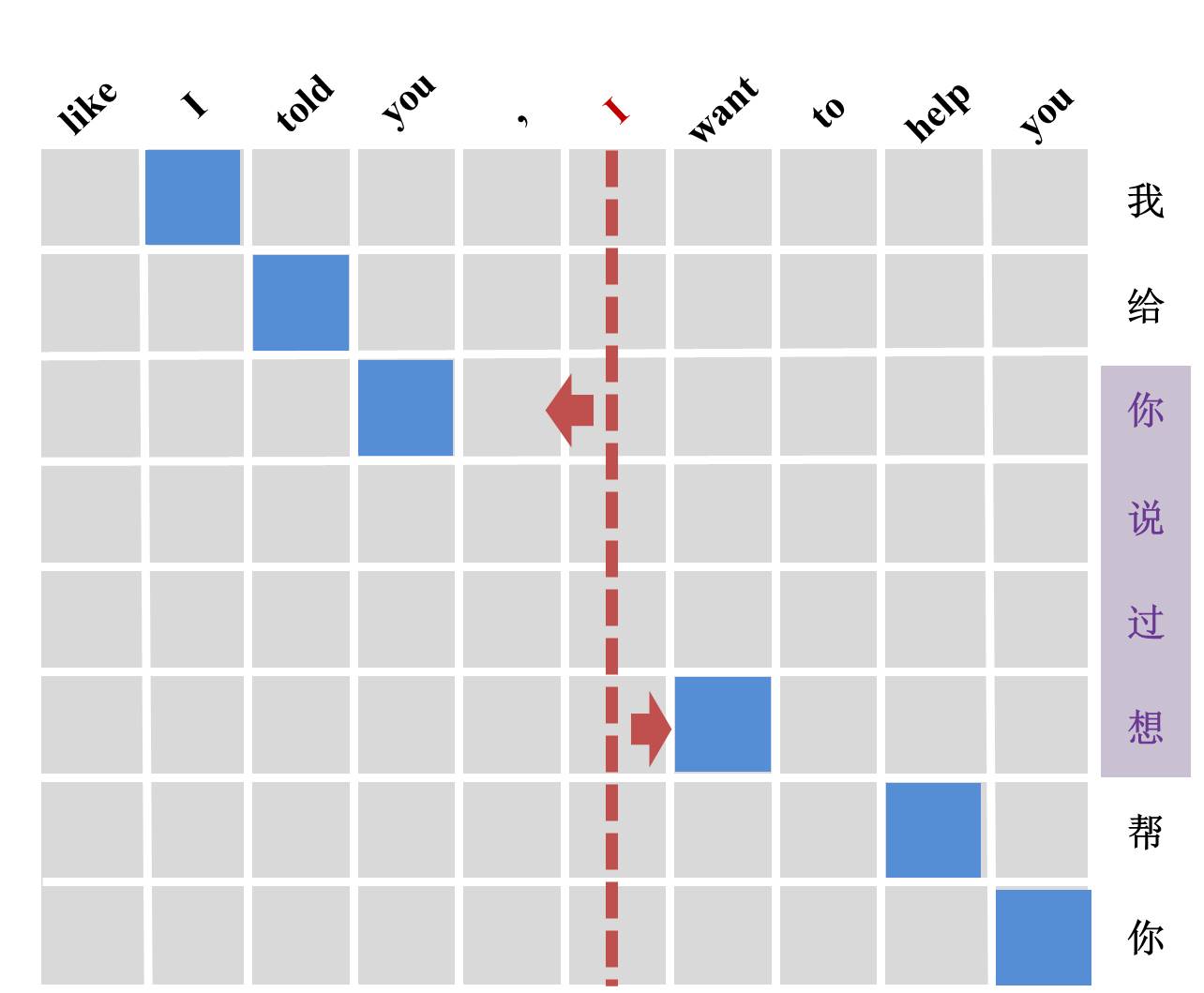}
\caption{Example of DP projection using alignment results (i.e. blue blocks).}
\label{fig:0}
\end{figure}

   
   
    

We use an example to illustrate our idea. Figure~\ref{fig:0} features a dropped pronoun ``\textit{我}'' (not shown) on the source side, which is aligned to the second ``{\em I}'' (in red) on the target side. 
For each pronoun on the target side (e.g. ``{\em I}'', ``{\em you}''), we first check whether it has an aligned pronoun on the source side. We find that the second ``{\em I}'' is not aligned to any source word and possibly corresponds to a $DP_{I}$ (e.g. ``\textit{我}'').
To determine the possible positions of $DP_{I}$ on the source side, we employ a {\em diagonal} heuristic based on the observation that there exists a diagonal rule in the local area of the alignment matrix. For example, the alignment blocks in Figure~\ref{fig:0} generally follow a diagonal line. Therefore, the pronoun "{\em I}" on the target side can be projected to the purple area (i.e. ``你 说 过 想'') on the source side, according to the preceding and following alignment blocks (i.e. ``\textit{you}-\textit{你}'' and ``\textit{want}-\textit{想}'').


However, there are still three possible positions to insert $DP_{I}$ (i.e. the three gaps in the purple area).
To further determine the exact position of $DP_{I}$, we generate possible sentences by inserting the corresponding Chinese DPs\footnote{The Chinese DP can be determined by using its English pronouns according to Table~\ref{tab:1}.  Note that some English pronouns may correspond to different Chinese pronouns, such as ``\textit{they} - 他们 / 她们 / 它们''. In such cases, we use all the corresponding Chinese pronouns as the candidates.} into every possible position. Then we employ an $n$-gram language model (LM) to score these candidates and select the one with the lowest perplexity as final result. 
This LM-based projection is based on the observation that the amount and type of DPs are very different in different genres.
We hypothesize that the DP position can be determined by utilizing the inconsistency of DPs in different domains. Therefore, the LM is trained on a large amount of webpage data (detailed in Section 3.1). Considering the problem of incorrect DP insertion caused by incorrect alignment, we add the original sentence into the LM scoring to reduce impossible insertions (noise).
\end{CJK}


\subsection{DP Generation}
\begin{CJK}{UTF8}{gbsn}

In light of the recent success of applying deep neural network technologies in natural language processing~\cite{Raymond:2007:INTERSPEECH,Mesnil:2013:INTERSPEECH}, we propose a neural network-based DP generator via the DP-inserted corpus (Section 2.1). We first employ an RNN to predict the DP position, and then train a classifier using multilayer perceptrons to generate our $N$-best DP results.

\subsubsection{DP detection}
The task of DP position detection is to label words if there are pronouns missing before the words, which can intuitively be regarded as a sequence labelling problem. We expect the output to be a sequence of labels $y^{(1:n)} = (y^{(1)}, y^{(2)}, \cdots, y^{(t)}, \cdots, y^{(n)})$ given a sentence consisting of words $w^{(1:n)} = (w^{(1)}, w^{(2)}, \cdots, w^{(t)}, \cdots, w^{(n)})$, where $y^{(t)}$ is the label of word $w^{(t)}$. In our task, there are two labels $L = \{NA, DP\}$ (corresponding to non-pro-drop or pro-drop pronouns), thus $y^{(t)} \in L$.

Word embeddings~\cite{mikolov2013distributed} are used for our generation models: given a word $w^{(t)}$, we try to produce an embedding representation $\mathbf{v}^{(t)} \in \mathbb{R}^d$ where $d$ is the dimension of the representation vectors. In order to capture short-term temporal dependencies, we feed the RNN unit a window of context, as in Equation (1):
\begin{equation}
\mathbf{x_d}^{(t)} = \mathbf{v}^{(t-k)} \oplus \cdots \oplus \mathbf{v}^{(t)} \oplus \cdots \oplus \mathbf{v}^{(t+k)}
\end{equation}
where $k$ is the window size.

We employ an RNN~\cite{Mesnil:2013:INTERSPEECH} to learn the dependency of sentences, which can be formulated as Equation (2):
\begin{equation}
\mathbf{h}^{(t)} = f(\mathbf{U}\mathbf{x_d}^{(t)} + \mathbf{V}\mathbf{h}^{(t-1)})
\end{equation}
where $f(x)$ is a sigmoid function at the hidden layer. $\mathbf{U}$ is the weight matrix between the raw input and the hidden nodes, and $\mathbf{V}$ is the weight matrix between the context nodes and the hidden nodes. At the output layer, a softmax function is adopted for labelling, as in Equation (3):
\begin{equation}
y^{(t)} = g(\mathbf{W_d}\mathbf{h}^{(t)})
\end{equation}
where $g(z_{m}) = \frac{e^{z_{m}}}{\sum_{k} e^{z_{k}}}$, and $\mathbf{W_d}$ is the output weight matrix.

\subsubsection{DP prediction}

\begin{table}[t]
\centering
\begin{tabular}{ c | l }
\hline
\bf ID. & \bf Description\\
\hline
\multicolumn{2}{c}{\bf Lexical Feature Set}  \\\hline
1 & $S$ surrounding words around $p$ \\
2 & $S$ surrounding POS tags around $p$ \\
3 & preceding pronoun in the same sentence\\
4 & following pronoun in the same sentence\\\hline
\multicolumn{2}{c}{ \bf Context Feature Set} \\\hline
5 & pronouns in preceding $X$ sentences\\
6 & pronouns in following $X$ sentences\\
7 & nouns in preceding $Y$ sentences\\
8 & nouns in following $Y$ sentences\\\hline
\multicolumn{2}{c}{ \bf Syntax Feature Set} \\\hline
9 & path from current word ($p$) to the root\\
10 & path from preceding word ($p-1$) to the root\\
\hline
\end{tabular}
\caption{List of features.}\label{tab:2} 
\end{table}

Once the DP position is detected, the next step is to determine which pronoun should be inserted based on this result. Accordingly, we train a 22-class classifier, where each class refers to a distinct Chinese pronoun in Table~\ref{tab:1}. We select a number of features based on previous work~\cite{xiang2013enlisting,yang2015recovering}, including lexical, contextual, and syntax features (as shown in Table~\ref{tab:2}). We set $p$ as the DP position, $S$ as the window size surrounding $p$, and $X,Y$ as the window size surrounding current sentence (the one contains $p$). For Features 1--4, we extract words, POS tags and pronouns around $p$. For Features 5--8, we also consider the pronouns and nouns between $X$/$Y$ surrounding sentences. For Features 9 and 10, in order to model the syntactic relation, we use a path feature, which is the combined tags of the sub-tree nodes from $p$/$(p-1)$ to the root. Note that Features 3--6 consider all pronouns that were not dropped. Each unique feature is treated as a word, and assigned a ``word embedding''. The embeddings of the features are then fed to the neural network. We fix the number of features for the variable-length features, where missing ones are tagged as None. Accordingly, all training instances share the same feature length. For the training data, we sample all DP instances from the corpus (annotated by the method in Section 2.1). During decoding, $p$ can be given by our DP detection model.

We employ a feed-forward neural network with four layers. The input $\mathbf{x_p}$ comprises the embeddings of the set of all possible feature indicator names. The middle two layers $\mathbf{a}^{(1)}$, $\mathbf{a}^{(2)}$ use Rectified Linear function $R$ as the activation function, as in Equation (4)--(5):
\begin{eqnarray}
\mathbf{a}^{(1)} & = & R(\mathbf{b}^{(1)}+\mathbf{W_p}^{(1)}\mathbf{x_p})\\
\mathbf{a}^{(2)} & = & R(\mathbf{b}^{(2)}+\mathbf{W_p}^{(2)}\mathbf{a}^{(1)})
\end{eqnarray}
where $\mathbf{W_p}^{(1)}$ and $\mathbf{b}^{(1)}$ are the weights and bias connecting the first hidden layer to second hidden layer; and so on. The last layer $\mathbf{y_p}$ adopts the softmax function $g$, as in Equation (6):
\begin{equation}
\mathbf{\mathbf{y_p}}  =  g(\mathbf{W_p}^{(3)}\mathbf{a}^{(2)})
\end{equation}


\end{CJK}



\subsection{Integration into Translation}

The baseline SMT system uses the parallel corpus and input sentences without inserting/generating DPs. As shown in Figure~\ref{fig:2},  the integration into SMT system is two fold: DP-inserted translation model ({\em DP-ins. TM}) and DP-generated input ({\em DP-gen. Input}).

\subsubsection{DP-inserted TM}
We train an additional translation model on the new parallel corpus, whose source side is inserted with DPs derived from the target side via the alignment matrix (Section 2.1). We hypothesize that DP insertion can help to obtain a better alignment, which can benefit translation. Then the whole translation process is based on the boosted translation model, i.e. with DPs inserted. As far as TM combination is concerned, we directly feed Moses the multiple phrase tables. The gain from the additional TM is mainly from complementary information about the recalled DPs from the annotated data.

\subsubsection{DP-generated input}
Another option is to pre-process the input sentence by inserting possible DPs with the DP generation model (Section 2.2) so that the DP-inserted input (Input ZH+DPs) is translated. The predicted DPs would be explicitly translated into the target language, so that the possibly missing pronouns in the translation might be recalled. This makes the input sentences and DP-inserted TM more consistent in terms of recalling DPs.

\subsubsection{N-best inputs} 
However, the above method suffers from a major drawback: it only uses the 1-best prediction result for decoding, which potentially introduces translation mistakes due to the propagation of prediction errors.
To alleviate this problem, an obvious solution is to offer more alternatives. Recent studies have shown that SMT systems can benefit from widening the annotation pipeline~\cite{Liu:2009,Tu:2010,Tu:2011,Liu:2013:ACL}.
In the same direction, we propose to feed the decoder $N$-best prediction results, which allows the system to arbitrate between multiple ambiguous hypotheses from upstream processing so that the best translation can be produced. 
The general method is to make the input with $N$-best DPs into a confusion network. In our experiment, each prediction result in the N-best list is assigned a weight of $1/N$.


\section{Related Work}



There is some work related to DP generation. One is zero pronoun resolution (ZP), which is a sub-direction of co-reference resolution (CR). The difference to our task is that ZP contains three steps (namely ZP detection, anaphoricity determination and co-reference link) whereas DP generation only contains the first two steps. Some researchers~\cite{Zhao:2007:EMNLP,Kong:2010:EMNLP,Chen:2013:EMNLP} propose rich features based on different machine-learning methods. For example,~\newcite{Chen:2013:EMNLP} propose an SVM classifier using 32 features including lexical, syntax and grammatical roles etc., which are very useful in the ZP task. However, most of their experiments are conducted on a small-scale corpus (i.e. OntoNotes)\footnote{It contains 144K coreference instances, but only 15\% of them are dropped subjects.} and performance drops correspondingly when using a system-parse tree compared to the gold standard one.~\newcite{Novak:2014:COLING} explore cross-language differences in pronoun behavior to affect the CR results. The experiment shows that bilingual feature sets are helpful to CR. Another line related to DP generation is using a wider range of empty categories (EC)~\cite{yang2010chasing,cai2011language,xue2013dependency}, which aims to recover long-distance dependencies, discontinuous constituents and certain dropped elements\footnote{EC includes trace markers, dropped pronoun, big PRO etc, while we focus only on dropped pronoun.} in phrase structure treebanks~\cite{xue2005penn}. This work mainly focus on sentence-internal characteristics as opposed to contextual information at the discourse level. More recently,~\newcite{yang2015recovering} explore DP recovery for Chinese text messages based on both lines of work. 

These methods can also be used for DP translation using SMT~\cite{chung2010effects,Nagard:2010:ACL,Taira:2012:SSSST,xiang2013enlisting}. \newcite{Taira:2012:SSSST} propose both simple rule-based and manual methods to add zero pronouns in the source side for Japanese--English translation. However, the BLEU scores of both systems are nearly identical, which indicates that only considering the source side and forcing the insertion of pronouns may be less principled than tackling the problem head on by integrating them into the SMT system itself. \newcite{Nagard:2010:ACL} present a method to aid English pronoun translation into French for SMT by integrating CR. Unfortunately, their results are not convincing due to the poor performance of the CR method~\cite{pradhan2012conll}. \newcite{chung2010effects} systematically examine the effects of EC on MT with three methods: pattern, CRF (which achieves best results) and parsing. The results show that this work can really improve the end translation even though the automatic prediction of EC is not highly accurate.

\section{Experiments}
\subsection{Setup}

For dialogue domain training data, we extract around 1M sentence pairs (movie or TV episode subtitles) from two subtitle websites.\footnote{Avaliable at \url{http://www.opensubtitles.org} and \url{http://weisheshou.com}.} We manually create both development and test data with DP annotation. Note that all sentences maintain their contextual information at the discourse level, which can be used for feature extraction in Section 2.1. The detailed statistics are listed in Table~\ref{tab:3}. As far as the DP training corpus is concerned, we annotate the Chinese side of the parallel data using the approach described in Section 2.1. There are two different language models for the DP annotation (Section 2.1) and translation tasks, respectively: one is trained on the 2.13TB Chinese Web Page Collection Corpus\footnote{Available at \url{http://www.sogou.com/labs/dl/t-e.html}.} while the other one is trained on all extracted 7M English subtitle data~\cite{wang2016lrec}. 

\begin{table}[h]
\centering
\begin{tabular}{m{1cm} m{0.8cm} m{1.5cm} m{1.5cm}m{0.8cm}}
\hline
Corpus & Lang. & Sentents & Pronouns & Ave. Len.  \\\hline
\multirow{2}{*}{Train} & ZH & 1,037,292 & 604,896 & 5.91 \\
  & EN & 1,037,292 & 816,610 & 7.87\\\hline
\multirow{2}{*}{Dev} & ZH & 1,086 & 756 & 6.13 \\ 
  & EN & 1,086 & 1,025 & 8.46\\\hline
\multirow{2}{*}{Test} & ZH & 1,154 & 762 & 5.81 \\
   & EN & 1,154 & 958 & 8.17\\\hline
\end{tabular}
\caption{Statistics of corpora.}\label{tab:3} 
\end{table}

We carry out our experiments using the phrase-based SMT model in Moses~\cite{Koehn:ACL:2007} on a Chinese--English dialogue translation task. Furthermore, we train $5$-gram language models using the SRI Language Toolkit~\cite{Stolcke:2002:CSLP}. To obtain a good word alignment, we run GIZA++~\cite{Och:2003} on the training data together with another larger parallel subtitle corpus that contains 6M sentence pairs.\footnote{Dual Subtitles -- Mandarin-English Subtitles Parallel Corpus, extracted by~\newcite{shikun2014} without contextual information at the discourse level.} We use minimum error rate training~\cite{Och:2003b} to optimize the feature weights. 

The RNN models are implemented using the common Theano neural network toolkit~\cite{Bergstra:2010:SCIPY}. We use a pre-trained word embedding via a lookup table. We use the following settings: windows = 5, the size of the single hidden layer = 200, iterations = 10, embeddings = 200. The MLP classifier use random initialized embeddings, with the following settings: the size of the single hidden layer = 200, embeddings = 100, iterations = 200.

For end-to-end evaluation, case-insensitive BLEU~\cite{Papineni:2002} is used to measure translation performance and micro-averaged F-score is used to measure DP generation quality. 

\begin{CJK}{UTF8}{gbsn}

\subsection{Evaluation of DP Generation}

We first check whether our DP annotation strategy is reasonable. To this end, we follow the strategy to automatically and manually label the source sides of the development and test data with their target sides. The agreement between automatic labels and manual labels on DP prediction are 94\% and 95\% on development and test data and on DP generation are 92\% and 92\%, respectively. This indicates that the automatic annotation strategy is relatively trustworthy.

We then measure the accuracy (in terms of words) of our generation models in two phases. ``DP Detection'' shows the performance of our sequence-labelling model based on RNN. We only consider the tag for each word (pro-drop or not pro-drop before the current word), without considering the exact pronoun for DPs. ``DP Prediction'' shows the performance of the MLP classifier in determining the exact DP based on detection. Thus we consider both the detected and predicted pronouns. Table~\ref{tab:4} lists the results of the above DP generation approaches. 
The F1 score of ``DP Detection'' achieves 88\% and 86\% on the Dev and Test set, respectively. However, it has lower F1 scores of 66\% and 65\% for the final pronoun generation (``DP Prediction'') on the development and test data, respectively. This indicates that predicting the exact DP in Chinese is a really difficult task. Even though the DP prediction is not highly accurate, we still hypothesize that the DP generation models are reliable enough to be used for end-to-end machine translation. Note that we only show the results of 1-best DP generation here, but in the translation task, we use $N$-best generation candidates to recall more DPs. 
\end{CJK}

\begin{table}[t]
\centering
\begin{tabular}{lllll}
\hline
\bf DP & \bf Set & \bf P & \bf R & \bf F1  \\\hline
\multirow{2}{*}{DP Detection} & Dev & 0.88 & 0.84 & 0.86  \\
& Test & 0.88 & 0.87 & 0.88 \\\hline
\multirow{2}{*}{DP Prediction}  & Dev & 0.67 & 0.63 & 0.65  \\
& Test & 0.67 & 0.65 & 0.66 \\\hline
\end{tabular}
\caption{Evaluation of DP generation quality.}\label{tab:4}
\end{table}

\subsection{Evaluation of DP Translation}

In this section, we evaluate the end-to-end translation quality by integrating the DP generation results (Section 3.3). Table~\ref{tab:5} summaries the results of translation performance with different sources of DP information. ``Baseline'' uses the original input to feed the SMT system. ``+DP-ins. TM'' denotes using an additional translation model trained on the DP-inserted training corpus, while ``+DP-gen. Input N'' denotes further completing the input sentences with the $N$-best pronouns generated from the DP generation model. ``Oracle'' uses the input with manual (``Manual'') or automatic (``Auto'') insertion of DPs by considering the target set. Taking ``Auto Oracle'' for example, we annotate the DPs via alignment information (supposing the reference is available) using the technique described in Section 2.1.

\begin{table}[t]
\centering
\begin{tabular}{m{2.65cm} m{2.1cm} m{2.1cm}}
\hline
\bf Systems & \bf Dev Set & \bf Test set  \\\hline
Baseline  &  20.06 & 18.76\\\hline
+DP-ins. TM & 20.32 (+0.26) & 19.37 (+0.61) \\
+DP-gen. Input \\
\qquad\qquad 1-best  & 20.49 (+0.43) & 19.50 (+0.74) \\ 
\qquad\qquad 2-best & 20.15 (+0.09) & 18.89 (+0.13) \\
\qquad\qquad 4-best & 20.64 (+0.58) & 19.68 (+0.92) \\
\qquad\qquad 6-best & 21.61 (+1.55) & 20.34 (+1.58) \\
\qquad\qquad 8-best & 20.94 (+0.88) & 19.83 (+1.07) \\\hline
Manual Oracle  & 24.27 (+4.21) & 22.98 (+4.22) \\
Auto Oracle  & 23.10 (+3.04) & 21.93 (+3.17)\\\hline
\end{tabular}
\caption{Evaluation of DP translation quality.}\label{tab:5}
\end{table}

The baseline system uses the parallel corpus and input sentences without inserting/generating DPs. It achieves 20.06 and 18.76 in BLEU score on the development and test data, respectively. The BLEU scores are relatively low because 1) we have only one reference, and 2) dialogue machine translation is still a challenge for the current SMT approaches.

By using an additional translation model trained on the DP-inserted parallel corpus as described in Section 2.1, we improve the performance consistently on both development (+0.26) and test data (+0.61). This indicates that the inserted DPs are helpful for SMT. Thus, the gain in the ``+DP-ins TM'' is mainly from the improved alignment quality. 
 
We can further improve translation performance by completing the input sentences with our DP generation model as described in Section 2.2. We test $N$-best DP insertion to examine the performance, where $N=$\{1, 2, 4, 6, 8\}. Working together with ``DP-ins. TM'', 1-best generated input already achieves +0.43 and + 0.74 BLEU score improvements on development and test set, respectively. The consistency between the input sentences and the DP-inserted parallel corpus contributes most to these further improvements. As $N$ increases, the BLEU score grows, peaking at 21.61 and 20.34 BLEU points when $N$=6. Thus we achieve a final improvement of 1.55 and 1.58 BLEU points on the development and test data, respectively. However, when adding more DP candidates, the BLEU score decreases by 0.97 and 0.51. The reason for this may be that more DP candidates add more noise, which harms the translation quality.

The oracle system uses the input sentences with manually annotated DPs rather than ``DP-gen. Input''. The performance gap between ``Oracle'' and ``+DP-gen. Input'' shows that there is still a large space (+4.22 or +3.17) for further improvement for the DP generation model.


\section{Case Study}

\begin{CJK}{UTF8}{gbsn}
We select sample sentences from the test set to further analyse the effects of DP generation on translation. 

\begin{figure}[t]
\centering
\graphicspath{ {figures/} }
\includegraphics[width=0.45\textwidth]{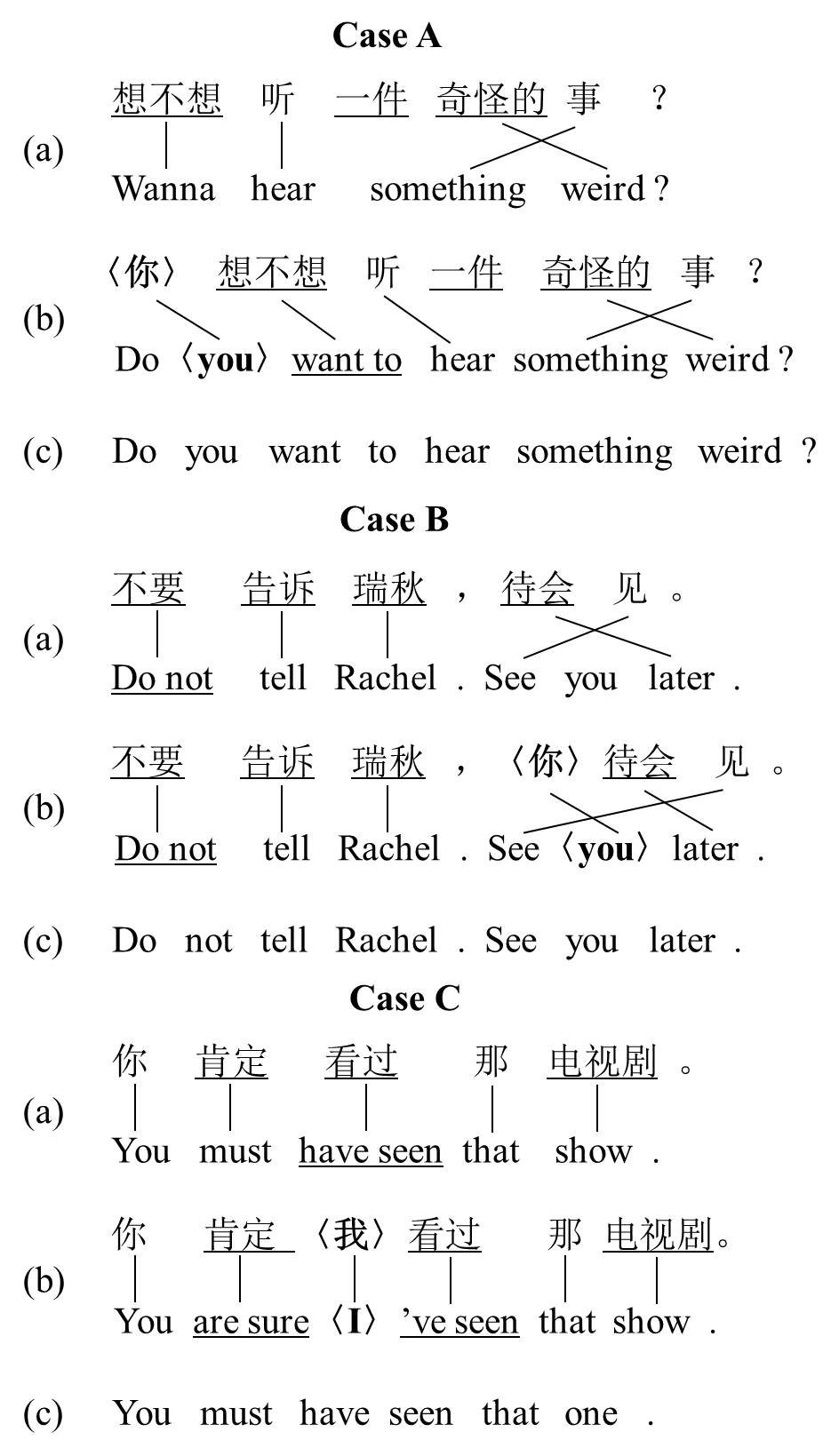}
\caption{Effects of DP generation for translation.}
\label{fig:4}
\end{figure}

In Figure~\ref{fig:4}, we show an improved case (Case A), an unchanged case (Case B), and a worse case (Case C) of translation no-/using DP insertion (i.e. ``+DP-gen. Input 1-best''). In each case, we give (a) the original Chinese sentence and its translation, (b) the DP-inserted Chinese sentence and its translation, and (c) the reference English sentence. In Case A, ``\textit{Do you}'' in the translation output is compensated by adding DP $\left<\text{你}\right>$ (you) in (b), which gives a better translation than in (a). In contrast, in case C, our DP generator regards the simple sentence as a compound sentence and insert a wrong pronoun $\left<\text{我}\right>$ (I) in (b), which causes an incorrect translation output (worse than (a)). This indicates that we need a highly accurate parse tree of the source sentences for more correct completion of the antecedent of the DPs. In Case B, the translation results are the same in (a) and (b). This kind of unchanged case always occurs in ``fixed'' linguistic chunks such as preposition phrases (``on \textit{my} way''), greetings (``see \textit{you} later'' , ``thank \textit{you}'') and interjections (``\textit{My} God''). However, the alignment of (b) is better than that of (a) in this case.
\end{CJK}

\begin{CJK}{UTF8}{gbsn}

\begin{figure}[h]
\centering
\graphicspath{ {figures/} }
\includegraphics[width=0.38\textwidth]{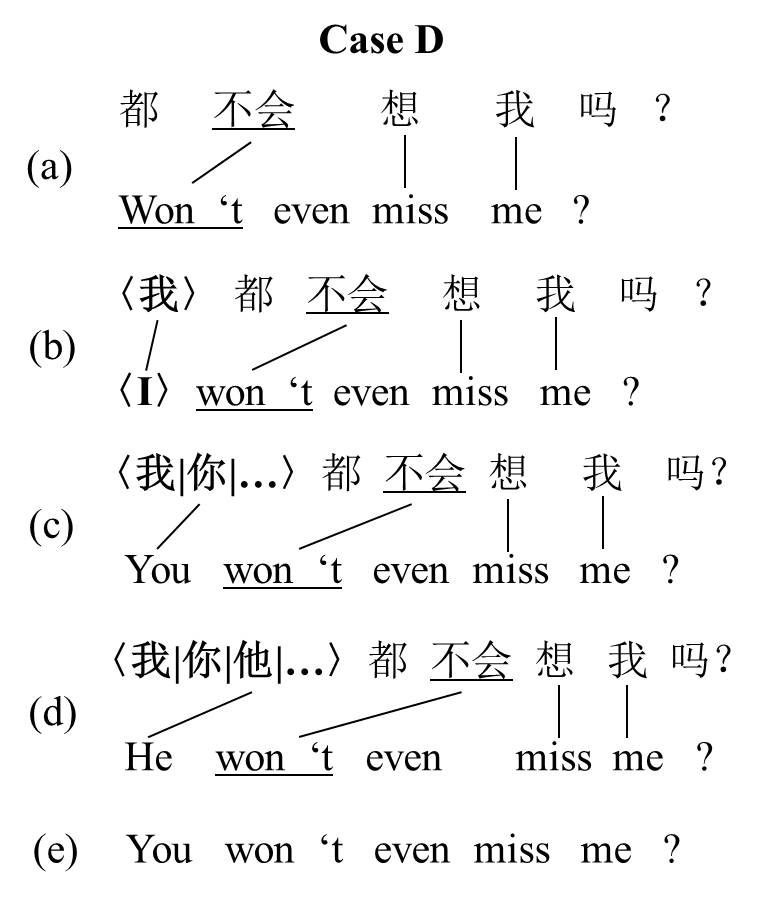}
\caption{Effects of N-best DP generation for translation.}
\label{fig:5}
\end{figure}

Figure~\ref{fig:5} shows an example of ``+DP-gen. Input N-best'' translation. Here, (a) is the original Chinese sentence and its translation; (b) is the 1-best DP-generated Chinese sentence and its MT output; (c) stands for 2-best, 4-best and 6-best DP-generated Chinese sentences and their MT outputs (which are all the same); (d) is the 8-best DP-generated Chinese sentence and its MT output; (e) is the reference. The $N$-best DP candidate list is $\left<\text{我}\right>$ (I), $\left<\text{你}\right>$ (You), $\left<\text{他}\right>$ (He), $\left<\text{我们}\right>$ (We), $\left<\text{他们}\right>$ (They), $\left<\text{你们}\right>$ (You), $\left<\text{它}\right>$ (It) and $\left<\text{她}\right>$ (She). In (b), when integrating an incorrect 1-best DP into MT, we obtain the wrong translation. However, in (c), when considering more DPs (2-/4-/6-best), the SMT system generates a perfect translation by weighting the DP candidates during decoding. When further increasing $N$ (8-best), (d) shows a wrong translation again due to increased noise. 

\end{CJK}

\section{Conclusion and Future Work}

We have presented a novel approach to recall missing pronouns for machine translation from a pro-drop language to a non-pro-drop language.
Experiments show that it is crucial to identify the DP to improve the overall translation performance. Our analysis shows that insertion of DPs affects the translation in a large extent. 

Our main findings in this paper are threefold:
\begin{itemize}[noitemsep,topsep=2pt]
\setlength\itemsep{0.3em}
\item Bilingual information can help to build monolingual models without any manually annotated training data;
\item Benefiting from representation learning, neural network-based models work well without complex feature engineering work;
\item $N$-best DP integration works better than 1-best insertion.
\end{itemize}

In future work, we plan to extend our work to different genres, languages and other kinds of dropped words to validate the robustness of our approach. 

\section*{Acknowledgments}

This work is supported by the Science Foundation of Ireland (SFI) ADAPT project (Grant No.:13/RC/2106), and partly supported by the DCU-Huawei Joint Project (Grant No.:201504032-A (DCU), YB2015090061 (Huawei)). It is partly supported by the Open Projects Program of National Laboratory of Pattern Recognition (Grant 201407353) and the Open Projects Program of Centre of Translation of GDUFS (Grant CTS201501).

\balance
\bibliography{naaclhlt2016}
\bibliographystyle{naaclhlt2016}

\end{document}